\documentclass[a4paper]{jpconf} 
\usepackage{graphicx}
\begin{document}
\title{Agricultural Plantation Classification using Transfer Learning Approach based on CNN}

\author{Uphar Singh, Tushar Musale, Ranjana Vyas, O.P.Vyas}

\address{Indian Institute of Information Technology, Allahabad, India}

\ead{pse2017003@iiita.ac.in}

\begin{abstract}
Hyper-spectral images are images captured from a satellite that gives spatial and spectral information of specific region.A Hyper-spectral image contains much more number of channels as compared to a RGB image, hence containing more information about entities within the image. It makes them well suited for the classification of objects in a snap. In the past years, the efficiency of hyper-spectral image recognition has increased significantly with deep learning. The Convolution Neural Network(CNN) and Multi-Layer Perceptron(MLP) has demonstrated to be an excellent process of classifying images. However, they suffer from the issues of long training time and requirement of large amounts of the labeled data, to achieve the expected outcome. These issues become more complex while dealing with hyper-spectral images. To decrease the training time and reduce the dependence on large labeled data-set, we propose using the method of transfer learning.The features learned by CNN and MLP models are then used by the transfer learning model to solve a new classification problem on an unseen data-set.  A detailed comparison of CNN and multiple MLP architectural models is performed, to determine an optimum architecture that suits best the objective. The results show that the scaling of layers not always leads to increase in accuracy but often leads to over-fitting, and also an increase in the training time.The training time is reduced to greater extent by applying the transfer learning approach rather than just approaching the problem by directly training a new model on large data-sets, without much affecting the accuracy.
\end{abstract}

\section{Introduction}
Remotely sensed data is the science and art of collecting knowledge in depth from an outsize distance, usually from satellites or aircraft, about various objects or areas.  Remotely sensed data is an integral aspect of data processing and interpretation, as the use of multiple wavebands in high resolution can be scanned over vast areas. It will differ by a few square feet of space. The quality in the key relies on the height or range at which the photographs were taken.[1]

Hyper-spectral devices collect aim images through a large variety of the electromagnetic spectrum. Hence, the hyper-spectral image (HSI) contains much simultaneous spectral and spatial detail. Because of its unique features, In many areas, HSI is widely used, including agriculture , medicine, and remote sensing. There is a large variety of thematic uses in contemporary society, such as biological research, geological research, hydro-logical science, and precision agriculture. [2, 3].

Hyper-spectral images of farm fields provide rich spectral-spatial information. This shared example offers greater forces to segregation. Zhen Ye (et al.) analyzes the application of spectral and spatial properties [14]. Every HSI reflects a slender wavelength in the electromagnetic spectrum, differing in the spectral unit.

It is possible to represent a hyper-spectral image as a 3-D data matrix: (P, Q, R). Here the square P and Q measure analogous to the image's 2 spatial dimensions, and the spectral dimension is analogous to R. The hyper-spectral sensor gets the luminosity or strength for a considerable range of slender spectral channel for any pixel in the Hyper-spectral image. Nevertheless, a hyper-spectral camera can catch an out-sized array, normally a whole bunch of thousands of contiguous spectral bands.

Hyper-spectral images range in scale from 10 to 22 nm and have slightly narrower bands. It's typically obtained from the partner spectrometer in imaging. Because of the high spectral detail in HSI, It gives them an even more greater ability to examine what's really unknown and make them unmistakably well suited for machine-driven image recognition and analysis. Therefore, the circulation of the hyper-spectral image makes it a great option to help sub-automate the deeply traditional method of classification of seed area distribution.They contain an excessive number of different choices that can be used by a broad array of Machine Learning algorithms.[7]

 CNN models are motivated by biology and impressed by study of D. H. Hubel and T. N. Wiesel. They projected proof of how primates interpreted the world all over each other utilizing complex brain neuron structure, and this successively motivated professionals to seek to establish analogous pattern recognition processes in deep learning. Advanced functional reactions produced by "complex cells" are formed by additional basic reactions by "simple cells" inside the visual cortex in their hypothesis.

CNN is a kind of neural network architecture,it allows one to extract representation of high image quality. Modern image recognition, where you recognize the properties of the image yourself, CNN takes raw pixel data from the image, trains the architecture, then, immediately draw the attributes for improved grouping.
 
 In 2005, the Information Processing Technology Office of the Defense Advanced Research Projects Agency Broad Agency Announcement (BAA) 05-29 offered a different transition training purpose: a system's capacity to understand and adapt information and expertise gained in preceding work to current work. Throughout this sense, transfer learning focuses at collecting the information through one or even more supply chain activities and applying information to a target task. Like multitask preparation, moving preparation worries more for the main task, instead of studying all of the source and goal activities simultaneously. The functions of root and goal activities in moving learning aren't more stellated.
 
 \begin{figure}[htbp]
\centerline{\includegraphics[scale=0.4]{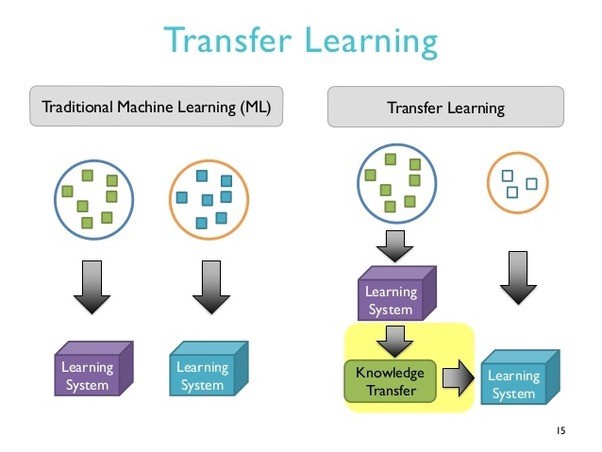}}
\caption{Traditional ML Vs Transfer Learning.}
\label{fig}
\end{figure}
Fig 1 demonstrates the gap amongst conventional learning methods and TL strategies. As we're seeing, modern machine learning strategies target at learning from scratch for every function, When transferring learning strategies, the goal is to shift knowledge from some previous tasks to a new challenge before fewer basic training data is accessible.

 Standard learning is distinct and focused solely on the preparation of particular functions, databases, and models isolated on them. No information which can be moved from one model to another is maintained. You will acquire information (attributes, weights, etc.) from previously proposed model to train new models in transferring learning, and even face challenges such as getting fewer data for new tasks.The organization of the manuscript is as follows: Section 2 states the literature Review part, section 3 is about the proposed methodology with pre-processing part, CNN Model training and Transfer learning followed by section 4 that explains results with dataset. Finally section 5 are about conclusion and future scope of the manuscript.
\\

\section{Literature Review}

In order to obtain spectral and spatial data for classification, hyperspectral image classification was mainly focused on some of the current understanding in previous studies. The scale, orientation and comparison of the spatial structures in the image involve the use of morphological operations to obtain features. [8]. An significant number of classification strategies for HSI have been developed over the past several years. Throughout this sense, several supervised learning approaches have been researched and updated for HSI classification in the machine learning group. Sparse representation suggested as spatial features or patch-based sparse representation [9]. The data characteristics modified from singular-pixel to neighborhood will collect richer details as spectral characteristics. Utilizing spectral-spatial training knowledge and classifying hyperspectral image data using traditional machine learning techniques such as k-nearest-neighbors[10], supporting vector machines[11], random forests (RFs), etc. Nevertheless, such approaches also involve a good previous knowledge of HSI, and the extraction phase is more disturbing and quick to miss essential characteristics.

An systematic analysis of the literature was undertaken on the various methodologies used to accomplish the proposed goal. Linear Discriminant Analysis shall be applicable to elements with lower strength of the Principal Component Analysis. These lower operating power modules with increased knowledge should not be thrown away, and can therefore be retrieved rather than classification. With this approach, classification performance is enhanced[4]. Two forms of Hyperspectral image classification techniques are commonly available: spectral classification models and spectral – spatial classification models[22]. Only spectral data collected from a hyperspectral detector will be used for the classification of spectral classification models. By contrast, spectral – spatial categorization approaches integrate the spatial data across the pixel to be classified.[23]

In order to achieve state-of-the-art precision, various deep learning models can be used to differentiate hyperspectral images that depend on both the spectral and spatial characteristics of the hyperspectral signal. Shutao Li et al provided a detailed comparative study of the specific deep learning frameworks used for hyperspectral identification including Deep Belief Networks(DBN), Generative Adversarial Networks(GAN's), Recurrent Neural Networks(RNNs) and Convolutionary Neural Networks(CNNs) and numerous variants such as the Gabor-CNN, S-CNN and RESNET.[29]

The paper by Ke Li et al. [34] explores another transfer learning approach for hyper-spectral image classification using deep belief networks wherein a limited Boltzmann machine Network is trained on the source domain data and its first few layers are extracted to be used for the target domain network. The target domain network is fine tuned further and used for classification of images in the target domain. The number of layers to be transferred are also varied and chosen for best accuracy.

It makes use of the Markov image feature to separate artifacts with class tags and prepare the convolution neural network on distinct channle findings from the datasets. Multiple 3-dimensional convolutionary filters of varying shapes are added simultaneously to the spatial deep convolution neural network. The convolution filters are applied, resulting in a combination of the spatial and spectral feature maps. The correct mark for every pixel of the image is successfully identified using these methodologies. This was checked on two datasets which were publicly accessible: the Indian Pines data collection and the Pavia University data collection. Comparative analysis of performances showed improved classification.[13]

It was found from the review of the literature that the convoluton neural network does all extraction and classification functionality. A 3-D kernel can also be used to leverage the spectral as well as spatial properties of hyperspectral images.

Hyperspectral Image is an image captured in such a way that each pixel includes complete spectrum. Thus the Hyperspectral Image forms a three dimensional data (i.e. is in the data cube format). In the Hyperspectral Dataset, we have an image along with its correct labels, also known as ground truth values.

Since the availability of the labeled Hyperspectral data is low, so to avoid overfitting and also to reduce the training time, we explore the transfer learning approach.Using related data from a different domain, we expand the knowledge for the problem domain.

The primary aim of this project is to classify plantations using Hyperspectral images. And we are also attempting to apply the Transfer Learning to build a scalable standardized framework.

\section{Proposed Methodology}
\subsection{Preprocessing}
Let the cube of spectral-spatial hyper spectral data be represented by $M \epsilon R^{P\times Q\times R}$ ,
where M is the representation of the initial input, P is the representation of width, Q is the height, and R is the quantity of spectral bands. Every hyperspectral image pixel in M comprise R spectral measurement and constitute a vector $Y = \left ( y_{1}, y_{2}...y_{C}  \right ) \epsilon R^{1\times 1\times C}$ , where C describes the land-cover classes.

Lots of bands with hyper spectral signal spectral band (network input) raise the expense of testing and prediction of computers System.

However, by means of a statistical methods of spectral responses of pixels belonging to a consistent framework, we can observe that the reaction deviation is exceedingly tiny. That implies that for each channel, pixels belonging to a common group have about the same values. Around the same moment, pixels belonging to a totally different group emit very different spectral characteristics. These features are assisted by the usage of a dimension reduction strategy to reduce the dimensional of input data in order to accelerate testing and estimation processes.[21]

First, the principal component analysis is implemented across the original hyper spectral image (M) through spectral channel to erase the spectral redundant information. The PCA decreases the quantity of spectral channels by R to B while preserving the equal spatial dimensions ( i.e. width P and height Q).

\begin{figure}[htbp]
\centerline{\includegraphics[scale=0.4]{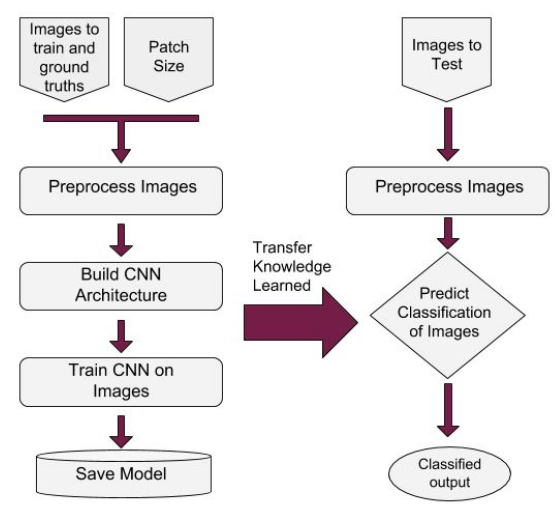}}
\caption{Framework of proposed method}
\label{fig}
\end{figure}

We just have decreased spectral channels that store local knowledge very necessary to classify any item. We potray PCA, The data cube is reduced by $X \epsilon R^{P\times Q\times B}$, where x is the updated input after PCA, P is the width, Q height and B is the quantity of spectral channels after PCA.

The hyper spectral image data cube is split into tiny intersecting 3D-patches to use the picture labeling methods, the truth labeling of which are determined by the centered pixel mark. We have built the adjacent 3D patches $P \epsilon R^{S\times S\times B}$, centered at the spatial position $\left ( a , b  \right )$, including the $S \times S$ frame or spatial extent and all B spectral channels.

\subsection{CNN Model Training}

Now create our model. The model we used contains three convolution layer after that one flatten layer and after this flatten layer three fully connected
(Dense) layers.

A 3D convolution is achieved by transforming the 3D data into a 3D kernel. The feature vector of the convolution layer in the developed hyperspectral image data model is created utilizing the 3D kernel across numerous adjoining channle in the input layer. The convolution is extended only to the spatial aspect of convolution 2D CNN. At the other side, it is ideal for the HSI classification issue to collect the spectral details encrypted in several channle across the spatial details. 2D CNN can't accommodate the spectral details. In comparison, the 3D convolution neural network kernel will concurrently derive spectral and spatial characteristic description from hyperspectral image but at the expense of high computational difficulty. We use 3D CNN instead of 2D CNN to take advantage of the 3D CNN adaptive function learning capabilities.

\begin{figure}[htbp]
\centerline{\includegraphics[scale=0.4]{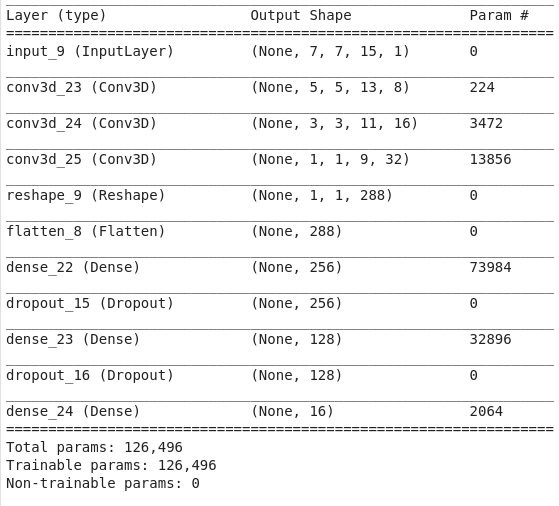}}
\caption{Summary of proposed model}
\label{fig}
\end{figure}

It included three 3D convolution layer, one flatten layer, and 3 fully connected layers. 
During this framework, the dimension of the 3D convolution kernel is  8 * 3 * 3 * 3 * 1, 16 * 3 * 3 * 3 * 8, and 32 * 3 * 3 * 3 * 16 Within the following 1st, 2nd and 3rd layers of convolution, respectively. Wherever 16 * 3 * 3 * 3 * 8 suggest that there are  16 3D kernels of dimension 3 * 3 * 3 for all eight 3D input characteristic maps. To extend at the same time the quantity of spectral-spatial feature maps, 3D convolution is applied thrice and then the flatten layer is applied. A complete understanding of the framework suggested shall be provided. This can be shown that inside the 1st dense layer the largest amount of parameters are available. The quantity of nodes in the previous compact layer is 16 and is the equal to the quantity of groups in the data collection for salinas. The cumulative variety of weight parameter trainable in this framework is 126,496. All these weights are configured and randomly prepared at random utilizing the Adam optimizer backpropagation method, using the softmax loss. We utilize Size 256 mini-batch and educate the network for 10 epochs.

After using this framework, the accuracy is 96 percent for window size of 7 with 10 epochs.

\subsection{Applying Transfer Learning}

The framework was prepare on Salinas Data and Tested on Salinas-A scene dataset. Testing dataset were firtst preprocessed.

During pre-processing, first the PCA is deployed across the test dataset along the spectral band to eliminate the spectral redundancy. We apply PCA on spectral channels, so that the amount of channle in the target images must equate the amount of spectral channels in the prepared images.

Similarally, what we have done in training phase, same we have to do in testing phase. The research dataset is split into tiny intersecting 3D blocks to use the classification method for the images, the truth table of which is determined by the center pixel label.

The weights obtained by training the CNN in the first phase were saved and later they are reloaded to train another model using Transfer learning. The CNN stump of the model i.e. Extract the convolutionary and pooling layers and the later Layers are removed, i.e. the fully connected layers and the softmax layers. To this convolutional stump, a new set of layers is connected with 3 fully connected layers with ReLU activation and a softmax layer of 6 units are appended. We are also using a 0.4 dropout for this with the fully connected layers. The layers excluding the fully connected layers and the softmax layers are frozen so that the weights for those layers are not modified and training only takes place for the fully connected layers and the softmax layer. The patches obtained from the Salinas-A scene dataset are then split into the training and the test set in a ratio of 40:60, and trained for 3 epochs. Architecture of the Transfer learning based model is summarized by Fig 4.
\begin{figure}[htbp]
\centerline{\includegraphics[scale=0.4]{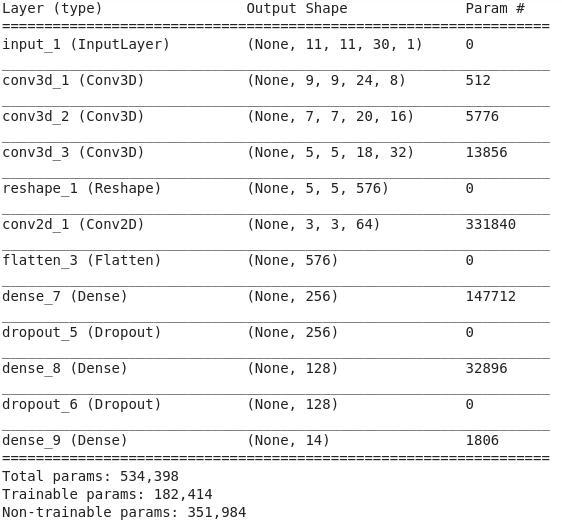}}
\caption{Summary of model built using Transfer Learning}
\label{fig}
\end{figure}

\section{RESULTS}
\subsection{Data set}
\subsubsection{Salinas scene (Training Data set)}
The scene was shot by Salinas Valley, California's 224-band Avery Censor, and features better spatial resolution (3.7-m pixels). The data set of Salina has 16 labels/categories, all of which show various crops, dissimilar types of land shield.

\begin{figure}[htbp]
\centerline{\includegraphics[scale=0.3]{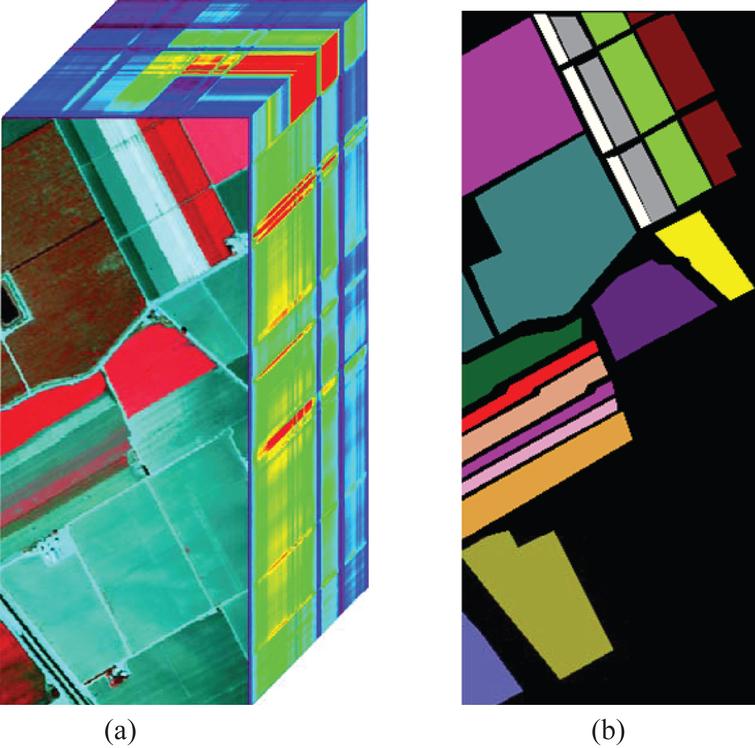}}
\caption{(a). Salinas Reference Band Dataset. (b). Salinas Data Set labels}
\label{fig}
\end{figure}

There are a total of 26,549 specimens of untrained grapes, earthen gardens. The total vineyards of grape samples were developed and trained without. it is made up 11,271 vineyard samples and 15,278 vineyard samples.

The table includes ground truth categories and their corresponding survey number for the Salinas scene.

\begin{table}[htbp]
\caption{Ground truth categories for the Salinas scene and their collections count}
\begin{center}
\begin{tabular}{|c|c|c|}
\hline
\textbf{Sr. no.} & \textbf{Label}               & \textbf{Count} \\ \hline
1           & Brocoli\_green\_weeds\_1     & 2009             \\ \hline
2           & Brocoli\_green\_weeds\_2     & 3726             \\ \hline
3           & Fallow                       & 1976             \\ \hline
4           & Fallow\_rough\_plow          & 1394             \\ \hline
5           & Fallow\_smooth               & 2678             \\ \hline
6           & Stubble                      & 3959             \\ \hline
7           & Celery                       & 3579             \\ \hline
8           & Grapes\_untrained            & 11271            \\ \hline
9           & Soil\_vinyard\_develop       & 6203             \\ \hline
10          & Corn\_senesced\_green\_weeds & 3278             \\ \hline
11          & Lettuce\_romaine\_4wk        & 1068             \\ \hline
12          & Lettuce\_romaine\_5wk        & 1927             \\ \hline
13          & Lettuce\_romaine\_6wk        & 916              \\ \hline
14          & Lettuce\_romaine\_7wk        & 1070             \\ \hline
15          & Vinyard\_untrained           & 7268             \\ \hline
16          & Vinyard\_vertical\_trellis   & 1807             \\ \hline
\end{tabular}
\label{tab1}
\end{center}
\end{table}

\subsubsection{Testing Dataset (Salinas-A scene)}
An tiny sub scene of Salinas pics, represented as Salinas-A. This contains 86 * 83 pixels and six sections.
As shown in fig 6 and fig 7.

\begin{figure}[h]
\begin{minipage}{15pc}
\includegraphics[width=15pc]{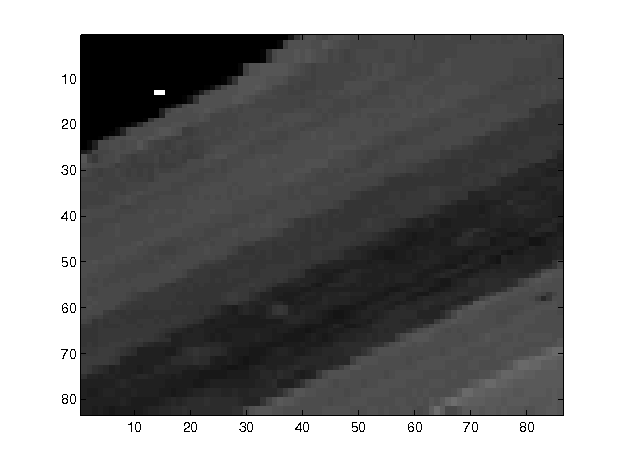}
\caption{Collection of channles for Salinas-A dataset}
\end{minipage}\hspace{2pc}
\begin{minipage}{15pc}
\includegraphics[width=15pc]{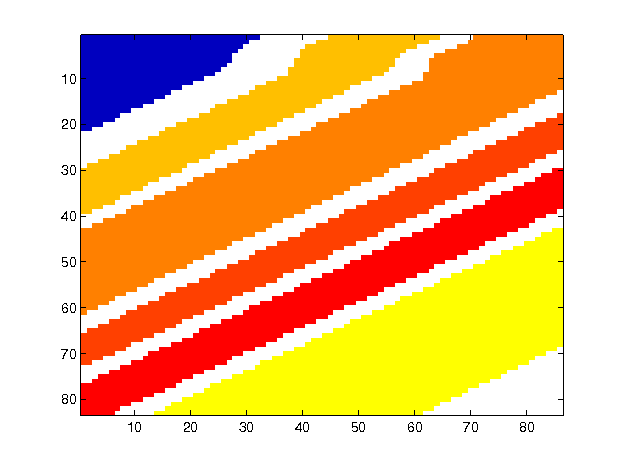}
\caption{Labels associates with Salinas-A dataset}
\end{minipage} 
\end{figure}

The Table II includes grades of ground reality for the Salinas-A scene and their corresponding survey number.

\begin{table}[htbp]
\caption{Ground truth categories for the Salinas-A scene and their collections count}
\begin{center}
\begin{tabular}{|c|c|c|}
\hline
\textbf{Sr. no} & \textbf{Label}               & \textbf{Count} \\ \hline
1           & Brocoli\_green\_weeds\_1     & 391              \\ \hline
2           & Corn\_senesced\_green\_weeds & 1343             \\ \hline
3           & Lettuce\_romaine\_4wk        & 616              \\ \hline
4           & Lettuce\_romaine\_5wk        & 1525             \\ \hline
5           & Lettuce\_romaine\_6wk        & 674              \\ \hline
6           & Lettuce\_romaine\_7wk        & 799              \\ \hline
\end{tabular}
\label{tab1}
\end{center}
\end{table}

\subsection{Results}
\subsubsection{Training Phase}
The basic simple convolution neural network model of one convolutionary layer,  one flatten layer and one dense layer was trained on Salinas dataset. The accuracy for this simple model is given in following table III.

\begin{table}[htbp]
\caption{Accuracy table for simple model}
\begin{center}
\begin{tabular}{|c|c|c|c|c|} 
\hline
\# & Patch Size & epoch & Spectral band & Accuracy  \\ 
\hline
1  & $3 \times 3$      & 10    & 15            & 84\%        \\ 
\hline
2  & $5 \times 5$      & 10    & 15            & 89\%        \\ 
\hline
3  & $7 \times 7$      & 10    & 15            & 93\%        \\
\hline
\end{tabular}
\label{tab1}
\end{center}
\end{table}

In the case of a 3 × 3, The output image consist of distinguished areas of Soil Vineyard, Untrained Vineyard and some other minor Grapes not trained collections as needed for a good categorization. However, the performance was poor in clarification relative to the greater output of the patch size and needed a larger patch size to resolve this.

In the occurrence of  a $7 \times 7$ pad, The clearness of the output image was considerably higher. This is because the $7 \times 7$ Pads could retrieve higher functionality than the $3 \times 3$ patch. The $7 \times 7$ Pad could find further information in the image.

It is obvious from the previous topic that higher pad sizes perform much better;As they can retrieve a greater depth of features.

Now we use a new CNN model which is discussed in the Methodology section. We train this model using the backpropagation method with the adam optimizer by utilizing the softmax loss. We use a mini-batch size of 256 and prepare the network for 10 iterations. The training dataset that is Salinas dataset is used in the CNN model.

The accuracy of this CNN model is $99.60 \%$.
\begin{frame}{}
    \begin{figure}[ht]
            \centering
            \includegraphics[width=20pc]{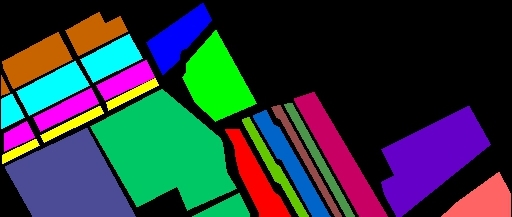}
            \caption{Groundtruth of CNN model}

        \hspace{0.5cm}
            \centering
            \includegraphics[width=20 pc]{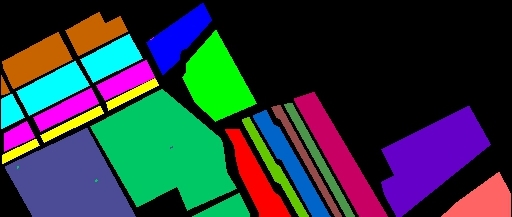}
            \caption{Prediction of CNN model}
    \end{figure}
\end{frame}

The groundtruth and predicted result for this model is given below in Fig 8 and Fig 9 respectively.

\subsubsection{Testing Phase}
For training over the Salinas-A scene dataset, the above model was loaded and the last few layers were popped off, as they were specifically trained on training Salinas dataset. New layers were added so that they could be trained specifically on the Salinas-A scene dataset.

We trained this model utilizing backpropagation method with the help of Adam optimizer by utilizing the softmax loss. We use mini batch size of 256 and prepare the model for 10 iterations.The accuracy for this is $99.8 \%$

The groundtruth and predicted result after using Transfer Learning on Salinas-A scene is given below in fig 10 and fig 11 respectively.
\begin{frame}{}
    \begin{figure}[ht]
        \begin{minipage}[b]{0.45\linewidth}
            \centering
            \includegraphics[width=\textwidth]{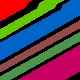}
            \caption{Groundtruth of Transfer Learning model}
            \label{fig:a}
        \end{minipage}
        \hspace{0.5cm}
        \begin{minipage}[b]{0.45\linewidth}
            \centering
            \includegraphics[width=\textwidth]{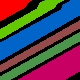}
            \caption{Prediction of Transfer Learning model}
            \label{fig:b}
        \end{minipage}
    \end{figure}
\end{frame}

\begin{table}[htbp]
\caption{Accuracy table for Both Datasets}
\begin{center}
\begin{tabular}{|c|c|c|c|c|} 
\hline
\  & CNN & CNN+Transfer Learning  \\ 
\hline
Accuracy      & 99.60    &  99.80               \\ 
\hline

\end{tabular}
\label{tab1}
\end{center}
\end{table}

\section{CONCLUSION \& FUTURE SCOPE}
\subsection{Conclusion}
The aim was to classify Hyperspectral Images of Sainas-A scene dataset. It was made the decision to use a supervised technique to learning with a training and test mode. Therefore, convolution neural network was selected to carry out the preparation assignment. The Salinas data-set, a publicly accessible data set with almost identical labels to the test data-set, has been used. The Transfer Learning method was used in which the convolution neural network architecture was developed utilizing Salinas data collection. The learned architecture was stored and the acquired information converted. This preserved architecture has been used to check the Salinas-A Sight data-set. Our model is shown to be effective across a number of architectures and data-sets, and thus has wide potential of application in hyper-spectral classification.

The advantages of Transfer Learning lies in the fact that we take advantage of the inner layer trained in the previous model and use its general information to train the outermost few layers. Thus, saving time by not having to train the inner layers again. The convolution neural network framework produced was developed on labels of Salinas data collection. Salinas field falsity groups include examples of several different crop forms. This pattern may therefore also be used for the classification of certain crops.

\subsection{Future scope}
As shown from the above studies the performance often improves with an change in patch size. A greater patch size will thus need even better processing capability. The usage of a computer with higher RAM or a high-performance computing machine would accommodate the even greater patch size.

The proposed methodology can also be improved to substitute the last fully connected layer with some other machine learning algorithm such as SVM.

\end{document}